\documentclass[10pt,twocolumn,letterpaper]{article}

\usepackage{cvpr}              

\usepackage{graphicx}
\usepackage{amsmath}
\usepackage{amssymb}
\usepackage{booktabs}
\usepackage[table,xcdraw]{xcolor}
\usepackage{subcaption}
\usepackage{comment}

\usepackage[pagebackref,breaklinks,colorlinks]{hyperref}

\newcommand{\pri}{\texttt{PRI} }
\newcommand{\post}{\texttt{POST} }

\usepackage[capitalize]{cleveref}
\crefname{section}{Sec.}{Secs.}
\Crefname{section}{Section}{Sections}
\Crefname{table}{Table}{Tables}
\crefname{table}{Tab.}{Tabs.}

\def\cvprPaperID{*****} 
\def\confName{CVPR}
\def\confYear{2023}

\begin{document}

\title{Lookism: The overlooked bias in computer vision} 

\author{Aditya Gulati\\
ELLIS Alicante\\
{\tt\small aditya@ellisalicante.org}
\and
Bruno Lepri\\
Fondazione Bruno Kessler\\
{\tt\small lepri@fbk.eu}
\and
Nuria Oliver\\
ELLIS Alicante\\
{\tt\small nuria@ellisalicante.org}
}
\maketitle

\begin{abstract}
   In recent years, there have been significant advancements in computer vision which have led to the widespread deployment of image recognition and generation systems in socially relevant applications, from hiring to security screening. However, the prevalence of biases within these systems has raised significant ethical and social concerns. The most extensively studied biases in this context are related to gender, race and age. Yet, other biases are equally pervasive and harmful, such as \emph{lookism}, \emph{i.e.} the preferential treatment of individuals based on their physical appearance. Lookism remains under-explored in computer vision but can have profound implications not only by perpetuating harmful societal stereotypes but also by undermining the fairness and inclusivity of AI technologies. Thus, this paper advocates for the systematic study of lookism as a critical bias in computer vision models. Through a comprehensive review of existing literature, we identify three areas of intersection between lookism and computer vision. We illustrate them by means of examples and a user study. We call for an interdisciplinary approach to address lookism, urging researchers, developers, and policymakers to prioritize the development of equitable computer vision systems that respect and reflect the diversity of human appearances.
\end{abstract}

\section{Introduction}
\label{sec:intro}

Computer vision systems are increasingly used to support decisions that impact critical aspects of people's lives, including hiring processes, security screening, social media interactions and healthcare diagnoses \cite{Gorbova2017, Ghadekar2023, Esteva2021}. Thus, there is a growing need to detect, quantify and mitigate the biases that such systems may perpetuate or even amplify \cite{Hall2022}. While there is significant work in the literature that has focused on gender \cite{Wang2019, Wang2020, Schwemmer2020}, racial \cite{Yucer2020,Howard2024,Khan2021} and age \cite{Karkkainen2021,JacquesJunior2019} biases, there is growing awareness of the existence of subtler biases that need to be accounted for \cite{Kumar2024}. \emph{Lookism} is one such bias. It consists of the preferential treatment of individuals based on their physical appearance. Rooted in societal standards of beauty and attractiveness and on our own cognitive biases \cite{Dion1972, Talamas2016, Eagly1991, Tversky1974}, lookism can lead to unequal treatment and reinforce harmful stereotypes when embedded in AI systems. The oversight of lookism as an important bias to consider in computer vision systems can result in systemic disadvantages and discrimination for individuals who do not conform to prevailing aesthetic norms, affecting their opportunities and how the are perceived and judged by automated systems. 

This paper aims to underscore the importance of studying and mitigating lookism within computer vision systems. By examining the roots and implications of this bias, we can develop strategies to ensure that computer vision algorithms promote fairness and inclusivity. 

\section{Lookism: A cognitive bias perspective}
\label{sec:lookismCogBias}

Lookism is deeply rooted in human cognitive biases \cite{Tversky1974}. From an evolutionary perspective, humans have developed a range of cognitive biases that influence our perceptions, memory and behaviors. Understanding lookism from the lens of these cognitive biases can shed light on how it manifests itself in computer vision systems. The most prominent cognitive biases that play a role in explaining lookism are briefly described next.

\textbf{1. The attractiveness halo effect} is a cognitive bias where the perception of one positive trait, \emph{i.e.} physical attractiveness, influences the perception of other unrelated traits, such as intelligence or moral character \cite{Dion1972,Kanazawa2011, Talamas2016}. When individuals are deemed attractive, they are often subconsciously perceived as more competent, sociable, and trustworthy \cite{frieze1991,Cash1985,Hamermesh1993,Miller1970,Todorov2008b}. This bias has been shown to lead to unfair advantages in various contexts, such as hiring, promotions, and social interactions \cite{Cash1985,frieze1991,Wilson2015}. 

\textbf{2. Aesthetics heuristics.} Heuristics are mental shortcuts that simplify decision-making \cite{Gigerenzer2009}. Aesthetic heuristics refer to the use of physical appearance as a fast and easy method to make judgments about a person’s other qualities. While these heuristics can be efficient, they often lead to oversimplified and biased evaluations that do not accurately reflect the individual's true abilities or characteristics \cite{Dion1972,Talamas2016}.

\textbf{3. The confirmation bias} is the tendency to search for, interpret, and remember information that confirms one’s preexisting beliefs or stereotypes \cite{Nickerson1998}. In the context of lookism, this bias could make people more likely to notice and remember positive behaviors from attractive individuals and overlook or rationalize negative behaviors. 

\section{Lookism and computer vision} The interplay between lookism and computer vision is two-fold. First, the advent of computer vision-based beauty filters could help mitigate the presence of this cognitive bias in humans by equalizing beauty. Second, similarly to humans, lookism might also be present in computer vision algorithms --leading to the concept of \emph{algorithmic lookism}-- from at least two perspectives. When computer vision systems are trained on datasets that reflect human biases, they can inadvertently learn and perpetuate lookism. For instance, if a facial recognition system is trained on images that disproportionately associate certain physical features with positive traits, it may develop a biased algorithm that favors those features. This can lead to unfair treatment and reinforce societal biases in automated decision-making processes and image generation systems. Furthermore, image generation and multimodal generative AI systems could be also impacted by lookism, leading to representational biases in the content they generate. 

In this section, we provide a brief overview of each of these aspects. 

\subsection{Beauty Filters and Lookism}
\label{sec:filtersAndLookism}

\begin{figure}
    \centering
    \includegraphics[width=0.8\linewidth]{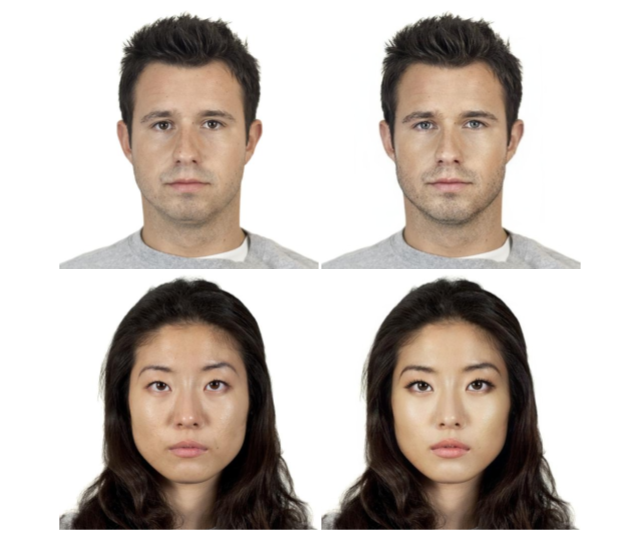}
    \caption{An example of images without (left) and with (right) beauty filters applied from our datasets.}
    \label{fig:imageSample}
\end{figure}

Beauty filters are a particularly popular family of computer vision based face filters which aim to \emph{beautify} the face of the person by automatically applying changes to the skin, the eyes and eyelashes, the nose, the chin, the cheekbones, and the lips. They rely on computer vision and augmented reality methods and their prevalence, with millions of users worldwide, profoundly impacts user self-presentation, raising questions about authenticity, self-esteem \cite{perrotta2020concept}, mental health \cite{cristel2021evaluation}, diversity \cite{Riccio2022} and racism \cite{Riccio2022racial}. 

Despite the issues raised by the use of beauty filters in daily life, they are a potentially powerful tool to study lookism in vision-based automated decision making systems since they enhance perceived attractiveness of individuals in images while preserving identity \cite{Riccio2022}. Additionally, beauty filters could be seen as a tool to mitigate lookism as they enable the democratization of beauty: used by millions of users to improve their appearances, particularly on social media and other digital spaces, beauty filters help level the playing field by allowing everyone to present themselves in ways that conform to societal beauty standards. This widespread accessibility could reduce the gap between those who naturally fit these standards and those who do not, potentially decreasing the social pressure and discrimination based on physical appearance.

To study the impact of beauty filters on lookism, we carried out a large-scale user study where participants provided their assessments of facial images from two datasets: 

The \textbf{\pri dataset}  consisting of a gender-balanced sample of 462 images from the Chicago Faces Database (CFD) \cite{Ma2015} and the FACES dataset \cite{Ebner2010}. These datasets have been used extensively for research on faces \cite{Pehlivanoglu2022,Holland2018,Wixted2017}. The CFD contains diversity in ethnicity but has limited diversity in age of the participants. The FACES dataset instead consists of images only of people who self-identified as white, but contains high diversity in age. Using images from both the CFD and FACES dataset together allows us to create a collection of images that has high age and ethnic diversity. Sample images from the \pri set can be seen on the left of Figure \ref{fig:imageSample}.

The \textbf{\post dataset} containing the images from the \pri dataset after applying a state-of-the-art beauty filter to them thereby creating a new set of \emph{beautified} images which depict the \emph{same} individuals as in the \pri set but in an ``attractive'' (or beautified) condition. Sample images from the \post set can be seen on the right of Figure \ref{fig:imageSample}.

We recruited $2,748$ participants from Prolific to rate the images on a 7-point Likert scale on perceived attractiveness and $6$ other attributes, including intelligence and trustworthiness, which have been shown to be impacted by lookism in the literature \cite{Batres2022,Oosterhof2008,Ma2015,Dion1972,Peterson2022,Golle2013,Peterson2022}. Each participant rated a balanced sample of $10$ images and were not told that half the images they see have a beauty filter applied to them. Thus, along with the two parallel datasets of high quality images, we obtained approximately 27,000 ratings by human annotators for attractiveness and 6 other dependent attributes.

Analysis of the ratings by human annotators revealed that beauty filters increased perceived attractiveness for all but 3\% of the images, for whom there was no change in perceived attractiveness. We also found that the age and gender of the person in the image played a significant role in perceptions of attractiveness with images of younger individuals and images of females receiving higher scores of attractiveness. Interestingly, ethnicity did not impact the attractiveness scores provided by the human annotators.

Furthermore, our analyses revealed that beauty filters could potentially mitigate the strength of lookism. Intelligence and trustworthiness exhibited weaker correlations with attractiveness in the \post set when compared to the \pri set. This however did not hold for other dependent attributes, such as sociability and happiness. Thus, the ability of beauty filters to reduce the spread of attractiveness could potentially be used to mitigate the strength of lookism in humans for some attributes, such as intelligence, but not for other attributes such as sociability. It is unknown, however if these results would apply to automated decision making systems that evaluate faces.

Furthermore, we found that beauty filters enhance dangerous gender stereotypes in society. While images of females received higher attractiveness scores than images of males, images of males were given significantly higher intelligence scores than those of females, with the gap between males and females increasing \emph{after beautification}. In other words, the application of beauty filters exacerbates existing gender biases, underscoring the need for a critical examination of these technologies \cite{Riccio2022, Riccio2024}. 

\subsection{Algorithmic Lookism}


Machine learning algorithms are typically trained on data which is annotated by humans. Thus, patterns of bias present in annotations provided by human raters are often present in these models \cite{Miceli2020,Chen2021}. A well-known example was in the hiring system deployed by Amazon which showed a strong bias against female employees \cite{Dastin2022}. 

We highlight below open questions associated with lookism in computer vision, followed by a discussion that highlights the challenges and ethical implications associated with studying lookism.

\subsubsection{Lookism in Image Generation}

\begin{figure*}
    \centering
    \begin{subfigure}{0.24\linewidth}
        \centering
        \includegraphics[width=\linewidth]{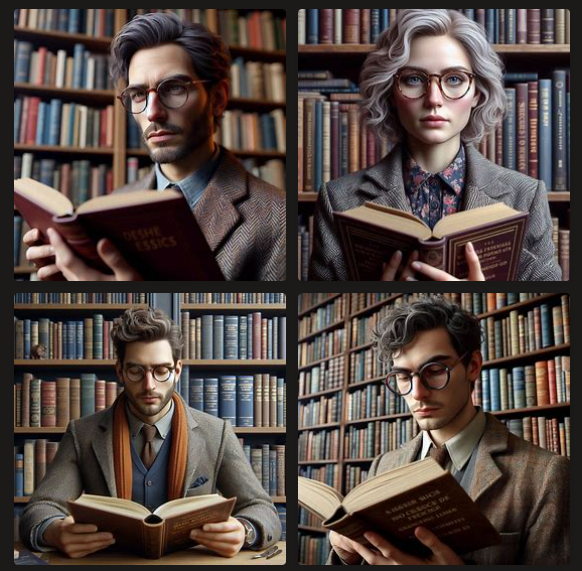}
        \caption{``An intelligent person''}
    \end{subfigure}
    \begin{subfigure}{0.24\linewidth}
        \centering
        \includegraphics[width=\linewidth]{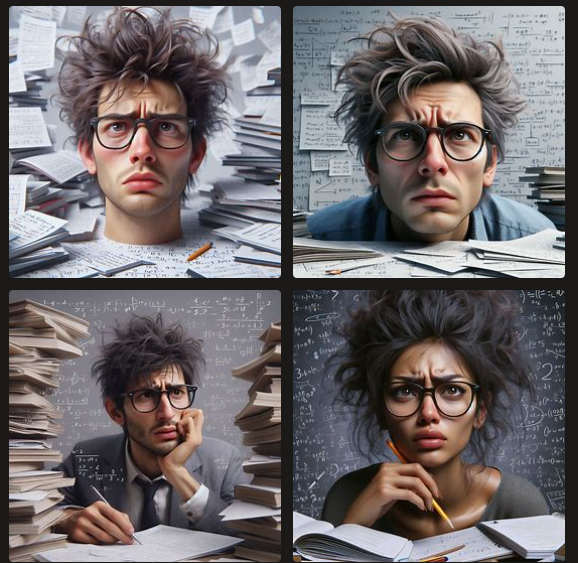}
        \caption{``An unintelligent person''}
    \end{subfigure}
    \centering
    \begin{subfigure}{0.233\linewidth}
        \centering
        \includegraphics[width=\linewidth]{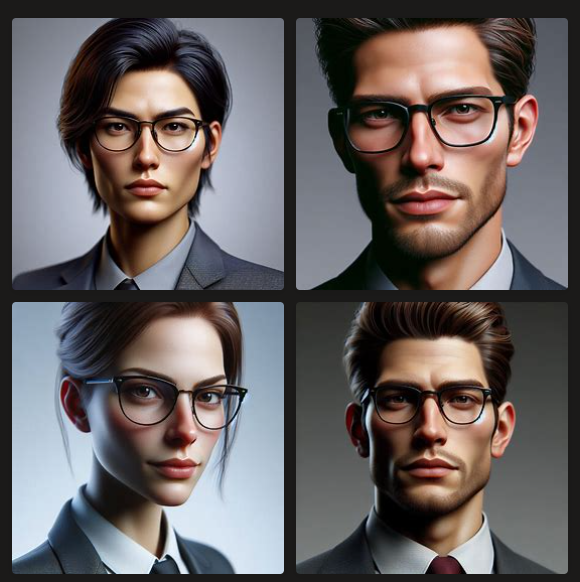}
        \caption{``A very competent person''}
    \end{subfigure}
    \begin{subfigure}{0.235\linewidth}
        \centering
        \includegraphics[width=\linewidth]{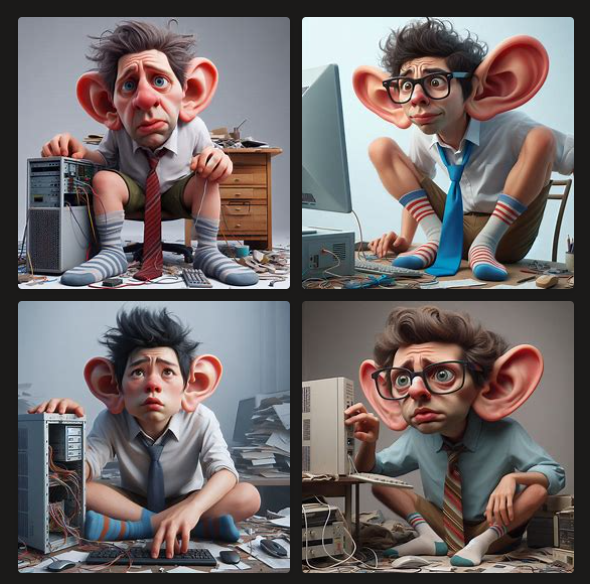}
        \caption{``A very incompetent person''}
    \end{subfigure}
    
    \caption{Illustrative examples of lookism in T2I models. Images generated by DALL-E3 through Microsoft's Copilot with prompts: ``create a hyperrealistic portrait of (a) an intelligent person; (b) an unintelligent person; (c) a very competent person; and (d) a very incompetent person''.}
    \label{fig:dalle}
\end{figure*}

Text-to-image (T2I) models are a class of machine learning models designed to generate images based on textual descriptions. They leverage recent advances in NLP and image generation methods to create a broad range of visual content, including representations of humans. The textual description is tokenized and projected to word or contextual embeddings from models like BERT \cite{Devlin2018}, GPT \cite{Achiam2023gpt} or more recent transformer-based architectures such as CLIP \cite{Radford2021_clip}. The image generation in early models was carried out using GANS \cite{Goodfellow2014} which have been superseded by variational autoencoders \cite{Kingma2013} and diffusion models \cite{Betker2023}. These systems typically include attention layers to help the model focus on specific parts of the text when generating the corresponding parts of the image, improving the alignment between the textual and the visual elements. 

While gender and racial biases have been studied in T2I models \cite{Naik2023}, little attention has been paid to the impact of lookism as a bias. AI-generated faces have been found to be perceived by humans as indistinguishable yet more trustworthy than faces of real people \cite{Nightingale2022}. Lookism would suggest that the reason these faces are more trustworthy is because they tend to be more attractive, yet an in-depth analysis to corroborate this hypothesis would be necessary. 

In fact, a systematic empirical study to unveil the presence of lookism in T2I systems would entail auditing them according to multiple dimensions by providing relevant prompts on topics, such as: (1) \emph{demographic representation}, involving the evaluation of how well the systems represent various ethnicities, genders, ages, body types and overall appreances in response to diverse prompts; (2) \emph{cultural and contextual sensitivity}, examining the system's ability to accurately and respectfully depict cultural symbols, attire, and settings, to assess to which degree the generated images perpetuate stereotypes or cultural insensitivity; (3) \emph{stereotype reinforcement} to investigate whether the T2I system amplify existing societal stereotypes, particularly regarding professions, personal attributes, social roles, and activities; (4) \emph{aesthetic diversity} to assess the range of visual styles and attractiveness standards the system produces; (5) \emph{realism and coherence} to focus on the technical quality of the generated images, evaluating whether the images are realistic and logically consistent with the provided descriptions; and (6) \emph{ambiguity vs specificity} to evaluate the system's performance with both highly specific and more ambiguous prompts, testing its ability to handle nuanced and complex descriptions. 

Figure \ref{fig:dalle} illustrates the existence of lookism in examples generated with DALL-E3 from Microsoft's Copilot when prompting the system to create images of intelligent vs unintelligent and competent vs incompetent persons. 

\subsubsection{Lookism in Decision-making Systems}

Gender and ethnicity-based biases have been studied extensively in computer vision systems that support human decisions in a variety of tasks, including emotion recognition \cite{Xu2020}, face recognition \cite{Robinson2020}, video surveillance \cite{Liu2019} and hiring \cite{Nguyen2015thesis,Kchling2020}. More recently, Multimodal Large Language Models (MLLMs) have also been evaluated for biases based on gender and ethnicity \cite{Kim2023, Booth2021}, but there is limited work evaluating these models for biases due to physical appearance.

Interestingly, a beauty bias has been reported in LLMs \cite{Kamruzzaman2023} which have exhibited significant positive correlations between attractiveness and personality traits (extraversion and conscientiousness) when automatically assessing personality from video transcriptions of job interviews \cite{Zhang2024}. 
Regarding MLLMs, Howard et al.\cite{Howard2024} have studied the impact of gender, race and physical appearance on predictions made by MLLMs by evaluating the description provided by these models on a large set of counterfactual image pairs \cite{Howard2023}. 

Given the scarcity of research on this topic, further work is needed to understand the extent to which lookism is present in vision-based decision-support models.

\section{Challenges}

The study of lookism is not exempt from challenges and ethical implications. 

First, attractiveness is a highly \emph{subjective} and \emph{cultural} construct. While the famous saying ``beauty is in the eye of the beholder'' suggests that perceptions of beauty vary significantly across individuals and cultures, there are studies that report an agreement across raters in perceptions of attractiveness \cite{Cunningham1995,Eisenthal2006,Perrett1998}, or at least in perceptions of unattractiveness \cite{Sorokowski2013}. 

Multimodal Large Language Models are unique when compared to other machine learning methods because of their ability to be used across multiple tasks. A potential solution to address the subjectivity of attractiveness would consist of asking the MLLM to evaluate attractiveness before proceeding with the desired task. However, preliminary experiments reveal that state-of-the-art MLLMs systems suffer from a \emph{positivity bias} and tend to assign extremely high scores to everyone, unlike the attractiveness scores given by human evaluators. Further research is needed here to understand how to evaluate the lookism bias multimodal systems. 

Second, there is a \emph{lack of awareness} of this bias and the \emph{inconsistency} of some of findings reported in the literature. Numerous studies on attractiveness have found that it is used as a cue for other unrelated human attributes \cite{Mitchem2015, Jackson1995}, such as perceived intelligence \cite{Batres2022,Talamas2016}. Yet, other studies have reported correlations between physical attractiveness and health \cite{Coetzee2009,Thornhill2006,Zebrowitz2004}, leading to doubts about treating lookism as a bias, even though these conflicting findings are reported in different contexts.

Third, the legal protection against lookism, or discrimination based on physical appearance significantly varies  across jurisdictions. Thus, lookism is not as widely recognized or legislated against as other forms of discrimination, such as those based on race, gender, age or disability \cite{Desir2010}. While there is a growing recognition of the need for more comprehensive laws and policies to address this bias, the legal protections against lookism remain inconsistent and limited. 

\section{Conclusion}

Computer vision systems that exhibit lookism can perpetuate and magnify societal biases, leading to the unequal treatment of individuals based on their looks. Furthermore, the deployment of certain computer vision apps and systems, such as beauty filters, raises concerns about the erosion of diversity and the perpetuation of narrow and \emph{white} beauty standards \cite{Riccio2022racial}. For these reasons, we believe that it is important for the computer vision community, in collaboration with experts of other domains, to devote efforts to detect, measure and mitigate lookism by ensuring diverse and representative training datasets, implementing fairness-aware algorithmic designs that consider lookism, and conducting continuous auditing and empirical evaluations to detect and rectify this bias. Addressing lookism is not only about preventing discrimination but also about fostering a more inclusive and equitable society where the algorithms that we design respect and reflect the rich diversity of human appearances. 

\section{Acknowledgments}

AG and NO are supported by a nominal grant received at the ELLIS Unit Alicante Foundation from the Regional Government of Valencia in Spain (Convenio Singular signed with Generalitat Valenciana, Conselleria de Innovacion, Industria, Comercio y Turismo, Direccion General de Innovacion), along with grants from the European Union’s Horizon 2020 research and innovation programme - ELISE (grant agreement 951847) and ELIAS (grant agreement  101120237), and by a grant from the Intel corporation. BL is partially supported by the European Union’s Horizon Europe research and innovation program under grant agreement No. 101120237 (ELIAS) and by the PNRR project FAIR - Future AI Research (PE00000013), under the NRRP MUR program funded by the NextGenerationEU.


\end{document}